\documentclass[10pt, conference, compsocconf]{IEEEtran}
\let\labelindent\relax

\usepackage{cite}
\usepackage{amsmath,amssymb,amsfonts}
\usepackage[symbol]{footmisc}
\usepackage{algorithmic}
\usepackage{graphicx}
\usepackage{textcomp}
\usepackage{graphicx}
\usepackage[square,sort,comma,numbers]{natbib}
\usepackage{caption}
\usepackage{float}
\usepackage{subcaption}
\usepackage{hyperref}
\usepackage{booktabs}
\usepackage{natbib}
\usepackage{enumitem}
\usepackage{xcolor}
\newtheorem{definition}{Definition}

\ifCLASSINFOpdf
\else
\fi
\begin{document}
%
\title{Bare Demo of IEEEtran.cls for IEEECS Conferences}


\makeatletter
\newcommand{\linebreakand}{%
  \end{@IEEEauthorhalign}
  \hfill\mbox{}\par
  \mbox{}\hfill\begin{@IEEEauthorhalign}
}
\makeatother

\def\BibTeX{{\rm B\kern-.05em{\sc i\kern-.025em b}\kern-.08em
    T\kern-.1667em\lower.7ex\hbox{E}\kern-.125emX}}

\title{Causality-Aware Next Location Prediction Framework based on Human Mobility Stratification
}

\author{
\IEEEauthorblockN{Xiaojie Yang\textsuperscript{*}}
\IEEEauthorblockA{
\textit{The University of Tokyo}\\
Tokyo, Japan \\
xiaojieyang@g.ecc.u-tokyo.ac.jp}\\

\IEEEauthorblockN{Takashi Michikata}
\IEEEauthorblockA{
\textit{The University of Tokyo}\\
Tokyo, Japan \\
takashi.michikata@koshizuka-lab.org}
\and
\IEEEauthorblockN{Zipei Fan\textsuperscript{*}}
\IEEEauthorblockA{
\textit{Jilin University}\\
Jilin, China \\
fanzipei@jlu.edu.cn}\\

\IEEEauthorblockN{Ryosuke Shibasaki}
\IEEEauthorblockA{
\textit{The University of Tokyo}\\
Tokyo, Japan \\
shiba@csis.u-tokyo.ac.jp}
\and
\IEEEauthorblockN{Hangli Ge}
\IEEEauthorblockA{
\textit{The University of Tokyo}\\
Tokyo, Japan \\
hangli.ge@koshizuka-lab.org}\\

\IEEEauthorblockN{Noboru Koshizuka}
\IEEEauthorblockA{
\textit{The University of Tokyo}\\
Tokyo, Japan \\
noboru@koshizuka-lab.org}
\thanks{\textsuperscript{*} Corresponding Authors.}

}


%


\maketitle

\begingroup\renewcommand\thefootnote{*}
\footnotetext{Corresponding Authors}
\endgroup

\begin{abstract}
Human mobility data are fused with multiple travel patterns and hidden spatiotemporal patterns are extracted by integrating user, location, and time information to improve next location prediction accuracy. In existing next location prediction methods, different causal relationships that result from patterns in human mobility data are ignored, which leads to confounding information that can have a negative effect on predictions. Therefore, this study introduces a causality-aware framework for next location prediction, focusing on human mobility stratification for travel patterns. In our research, a novel causal graph is developed that describes the relationships between various input variables. We use counterfactuals to enhance the indirect effects in our causal graph for specific travel patterns: \textit{non-anchor targeted travels}. The proposed framework is designed as a plug-and-play module that integrates multiple next location prediction paradigms. We tested our proposed framework using several state-of-the-art models and human mobility datasets, and the results reveal that the proposed module improves the prediction performance. In addition, we provide results from the ablation study and quantitative study to demonstrate the soundness of our causal graph and its ability to further enhance the interpretability of the current next location prediction models.
\end{abstract}


%
\IEEEpeerreviewmaketitle

\section{Introduction}
Next location prediction plays a crucial role in urban location-based services \cite{wu2018location}. In this technique, human mobility data (including GPS trajectories and check-in records) are used to predict future locations within a fixed time window. In human mobility data, \textbf{anchor} locations, such as \textit{home}, \textit{workplace}, and \textit{school} \cite{jiang2017activity}, of an individual, are used to model regular travel patterns. Predicting human mobility based on revisiting behaviors at these anchor locations is simple with statistical estimations \cite{song2010modelling}. However, this prediction task becomes challenging when dealing with \textbf{nonanchor} locations. Unlike anchor locations, travels to nonanchor locations are not part of a user’s daily routine and exhibit unpredictable patterns with limited periodicity. Therefore, prediction on nonanchor locations requires understanding human mobility and properly distinguishes these travels.
\begin{figure}[H]
\centering
    \begin{subfigure}[b]{0.3\textwidth}
        \includegraphics[height=1.4in, width=\linewidth]{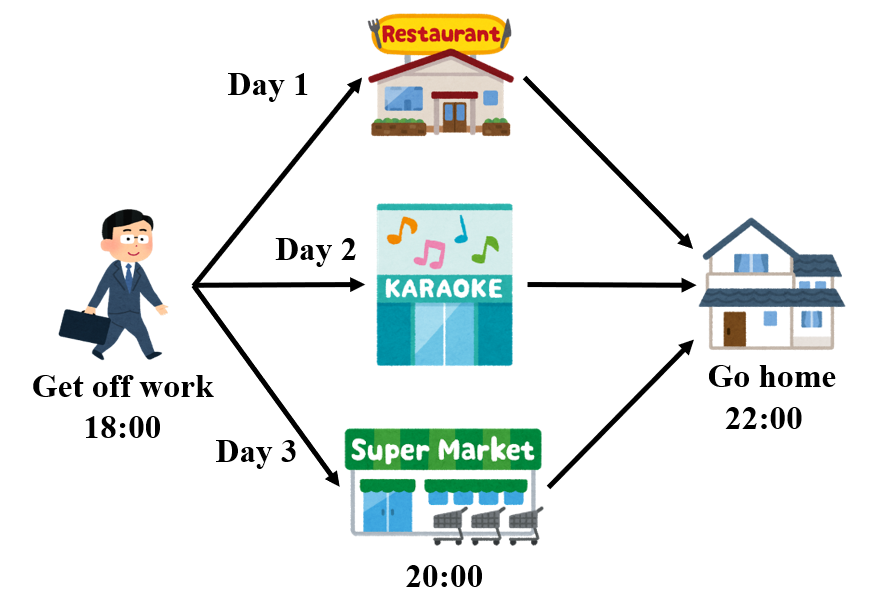}
        \subcaption{Anchor targeted}\label{anchor}
    \end{subfigure}
    \begin{subfigure}[b]{0.35\textwidth}
    \includegraphics[height=0.6in,width=\linewidth]{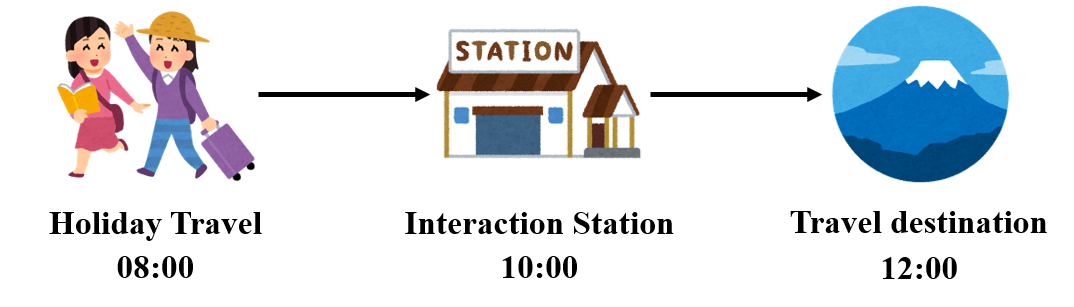}
        \subcaption{Nonanchor targeted}\label{nonanchor}
    \end{subfigure}
    \caption{Two travel patterns.}
    \label{examples}
\end{figure}
\indent As depicted in Fig. \ref{anchor}, a white-collar worker engaging in post-work activities before returning home is considered. In this scenario, mobility in three travels can be categorized as anchor-targeted because these travels eventually end at a common anchor location: \textit{home}. Using historical data, we can estimate the possibility of returning home at night \cite{zhao2016urban}. This approach is a conventional framework for next location predictions. However, in this method, previous locations may not be as informative as initially presumed, typically when travel ends at anchor locations. In contrast, previous location information becomes relatively more critical for non-anchor-targeted travels, such as those travels that occur during holidays or business trips (as shown in Fig. \ref{nonanchor}). Knowing that these two different travel patterns exist in human mobility data, it becomes important to stratify the data and impose different prediction strategies.
\indent To statistically analyze the differences between anchor and nonanchor-targeted travels and stress the necessity of data stratification, we use the Foursquare dataset \cite{yang2014modeling} in Fig. \ref{biased} to demonstrate the effectiveness of previous location information (e.g., a \textit{Mall}), in accurately estimating different destinations. For example, previous locations are not helpful despite having time information simultaneously for travels to anchor locations, such as \textit{Subway}. This conclusion aligns with the findings of previous studies \cite{gambs2012next}, whereas for nonanchor-targeted travels (e.g., to a \textit{Noodle House}), previous locations can considerably improve prediction performance at most times of the day. Based on these considerations, we stratify human mobility data based on individual level \textit{visit frequency} to various locations.
\begin{figure}[H] 
    \begin{subfigure}[b]{0.45\textwidth}
        \includegraphics[width=\linewidth]{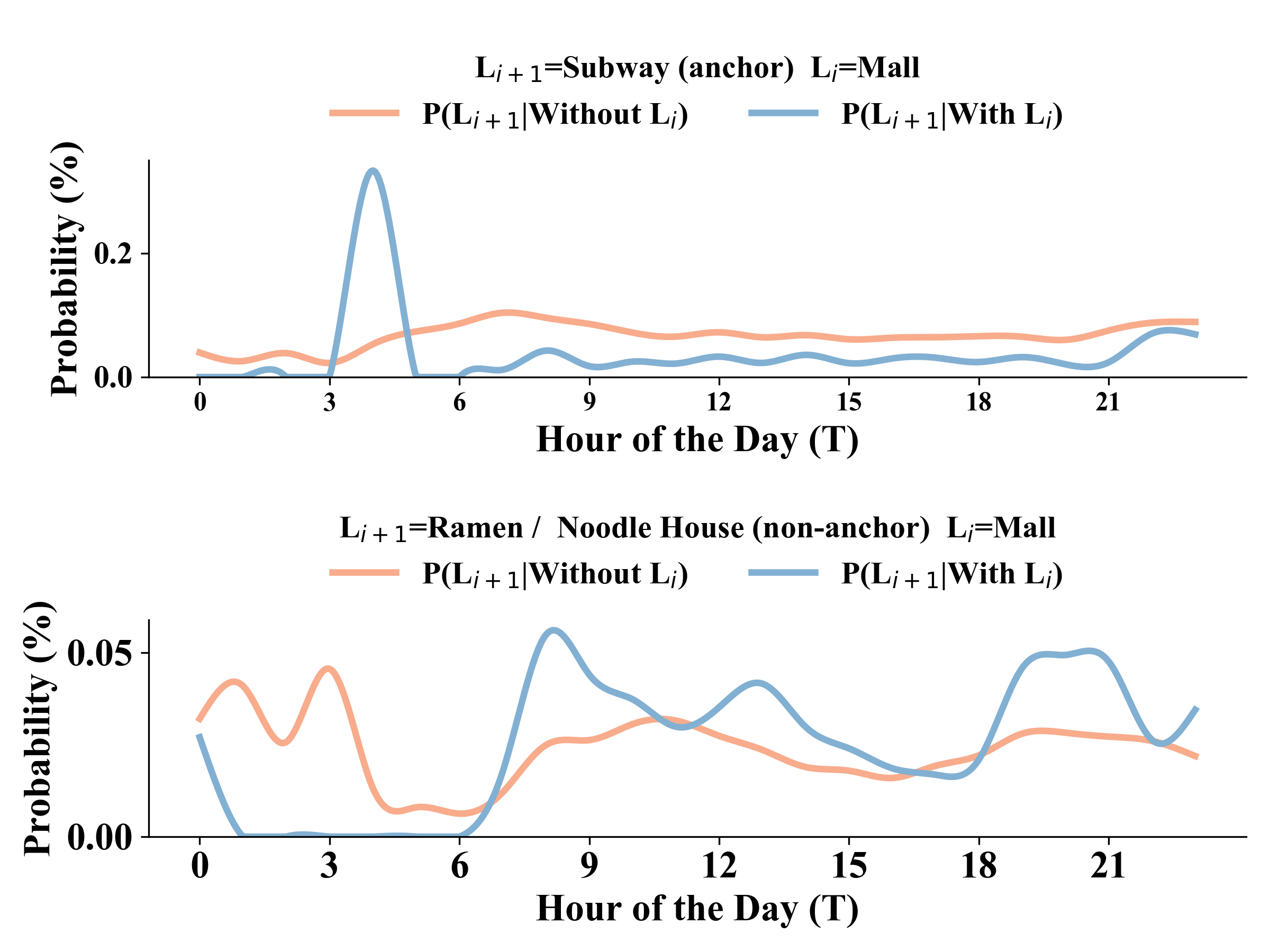}
    \end{subfigure}
    \caption{Statistical impacts of knowing previous location or not on the next location predictions.}
    \label{biased}
\end{figure}
\indent Here we want to introduce \textbf{\textit{causal inference}}, a framework that describes and measures causal relationship among variables during prediction. To explain causality in next location prediction, we incorporate \textit{causal inference} as a graphical model for a cause-effect interpretation \cite{verma1990causal}. Conventionally, previous research focused on integrating user, location and time information to create a hidden state, denoted $\textit{\textbf{H}}$ depicted in Fig. \ref{conventional_causal}. However, as we explained before, location information can be a \textbf{\textit{confounder}} (a variable that interference prediction process) for predicting outcomes $\textit{\textbf{Y}}$. To address this limitation, we introduce a new model represented in Fig. \ref{proposed_causal} to update the relationships during prediction. Two core principles guide our proposed causal graph: i) introduction of a new hidden state $\textit{\textbf{G}}$ as an intermediate variable; ii) establishment of two direct causal relationships: $\textit{\textbf{L}} \rightarrow \textit{\textbf{Y}}$ and $\textit{\textbf{H}} \rightarrow \textit{\textbf{Y}}$. With proposed causal graph, we can comprehensively describe a framework for next location prediction for anchor and non-anchor targeted travels after human mobility stratification and we will give more details in the methodology.\\
\indent The key contributions of this study are as follows:
\begin{itemize}[topsep=0pt]
\item[$\bullet$] We develop a graph-based causality framework that allows a comprehensive description of cause-effect relationships between variables for next location prediction.
\item[$\bullet$] We apply causal inference for nonanchor targeted travels after stratification to enhance the indirect causal effect during prediction with counterfactuals strategies.
\item[$\bullet$] The proposed framework is evaluated with several state-of-the-art models and human mobility datasets, and the results highlight a notable performance improvement.
\end{itemize}
\begin{figure}[htbp]
        \centering
        \subcaptionbox{\label{conventional_causal}}{
        \includegraphics[width = .35\linewidth]{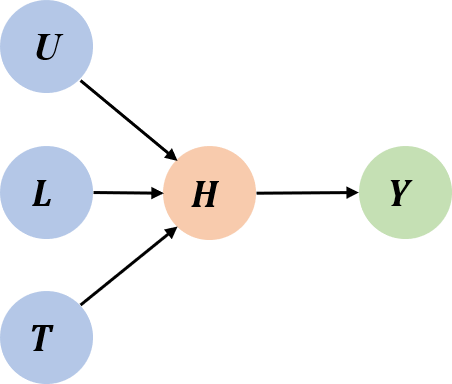}
    }
        \subcaptionbox{\label{proposed_causal}}{
        \centering
        \includegraphics[width = .35\linewidth]{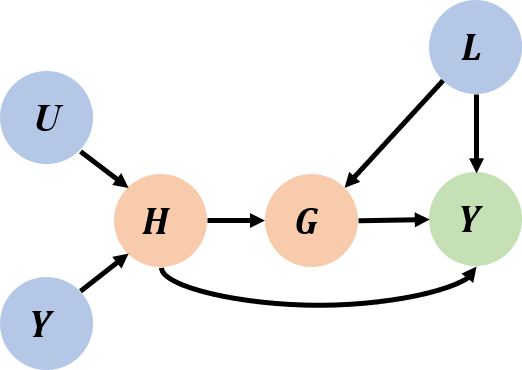}
    }
        \caption{Causal graph of next location prediction: (a) conventional, and (b) our approach.. Observations $\{U$, $L$, $T\}$ in (a) and (b) refer to user, location and time information; $\{H, G\}$:  hidden states; $Y$: next predicted location}
        \label{causal}
\end{figure}
\section{Related Work}\label{related}
\noindent \textbf{Next Location Prediction.} Predicting individuals' future locations can benefit multiple location-based services and applications like transportation management \cite{ge2024k, hangli2022multi}, public health monitoring, travel recommendation, and so on \cite{luca2021survey}. With the growth in the volume of human mobility data and computational power, Recurrent neural networks (RNNs) have emerged as a dominant method for uncovering naturally time-varying patterns hidden in individual-level human mobility \cite{islam2022survey}. Additionally, incorporating external factors such as semantic information of Points of Interests (POI) \cite{yao2017serm}, historical trajectories \cite{feng2018deepmove}, and spatiotemporal attention \cite{li2020hierarchical,luo2021stan}, the interpretability of this task has also been developed in depth. These previous studies considered travel records (trajectories) as sequential data, focusing on hidden feature extraction related to spatiotemporal transitions. Furthermore, the development of deep learning also resulted in the integration of techniques from other fields, such as Transformers \cite{yang2022getnext}, and graph structures \cite{wang2022graph,gao2022contextual}.\\
\noindent \textbf{Causal Inference.} With deep learning growing over the years, the reasoning capability of models is getting more and more attention \cite{scholkopf2021toward}. To answer causality relationships for interpretability, causal inference was introduced to provide a novel tool in data mining \cite{pearl2009causal}.And this methodology is increasingly widely applied in the rapidly evolving machine learning domain, enhancing estimation methods in areas such as advertising, recommendation, and medicine \cite{yao2021survey}. Related applications have extended to human mobility in recent years, for example, \cite{ma2022assessing} provided a causality analysis for different policies' effectiveness in controlling death due to the spread of COVID-19. As for extreme weather events, such as typhoons, causal inference aided in identifying and mitigating confounders affecting human mobility \cite{zhiwen2023assessing}.
\section{Preliminary}\label{formulation}
In this section, several critical definitions in this study are explained for next location prediction using human mobility data with causal inference.
\begin{definition}[\textbf{Human Mobility Data}]
\textup{
As mentioned, human mobility data mainly consist of user, location, and time information. Users are denoted as $u_{i}\in U$, where $U = \{u_{i} | i=1, 2, 3, ...\}$. The locations in our study are defined as spatial entities, encompassing (POI) and regions, and they are indexed and denoted as $L = \{l_{i} | i=1, 2, 3, ...\}$. Finally, the time information is simplified to $hour$ by referring to $T = \{t_{i} | i=1, 2, 3, ..., 24\}$.
}
\label{mobility data}
\end{definition}
\begin{definition}[\textbf{Anchor Locations}]
\textup{
To identify anchor locations, we count frequency of visits to various locations. A location becomes an individual-level anchor location of user $u_{i}$, $L^{anchor}_{i}$, when the frequency of visits exceeds a predefined threshold.
}
\label{anchor locations}
\end{definition}
\begin{definition}[\textbf{Human Mobility Stratification}]
\textup{
This study conceptualizes anchor and nonanchor locations and stratify human mobility data based on travel destinations. For user $u_{i}$, the $j$-th trajectory is represented by $traj_{i, j} = \{(u_{i}, l^{k}_{i, j}, t^{k}_{i, j}) | k = 1, 2, 3, ...\}$, where $k$ is the record index. This representation allows raw data to be organized into a processed trajectory dataset, denoted as $\mathcal{T}=\{traj_{i, j}\}$. For stratification, we differentiate anchor-targeted travels $\mathcal{T}^{1}$ and nonanchor-targeted travels $\mathcal{T}^{2}$, based on the individual-level anchor location set $L^{anchor}_{i}$.
}
\label{stratification}
\end{definition}
We propose a Causality-Aware Next Location Prediction framework. The task is to predict destinations on trajectories in $\mathcal{T}$. For $\mathcal{T}^{2}$, we integrate counterfactuals to optimize prediction performance. The objective is to improve the overall prediction performance with metrics that include recall, NDCG, and MRR compared to previous models \cite{wang2021libcity}.
\section{Methodology}
In this section, we introduce how to integrate causality into the next location prediction.
\subsection{Causal Inference Formulation}\label{causality}
In this section, we introduce the notations and formulations for causal analysis in human mobility. The causal graph is depicted as a directed acyclic graph (DAG) denoted as $G=(V, E)$. Here, $V$ represents the set of variables, and $E$ represents the directed edges that indicate the cause-effect relationships between these variables \cite{niu2021counterfactual}. In this study, the causal relationships are illustrated in Fig. \ref{proposed_causal}. We distinguish two causality paths: direct causality, represented as $H \rightarrow Y$ and $L \rightarrow Y$; indirect causality, depicted through the pathway $H \rightarrow G \rightarrow Y$. Each of these causality paths influences the final predicted outcomes with different effects \cite{tian2002general}.
\begin{figure}[htbp] 
\centering 
\includegraphics[height=1.5in,width=0.4\textwidth]{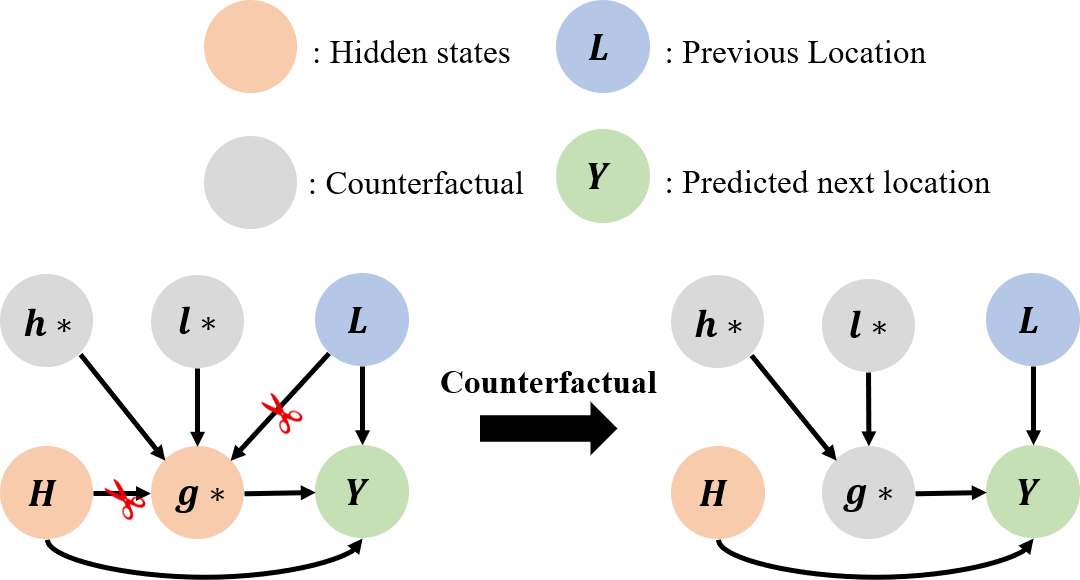} 
\caption{Calculating causal effects through counterfactual} 
\label{counterfactual} 
\end{figure}\\
\indent \textbf{Causal effects} compare the outcomes under varying treatment conditions \cite{rubin1978bayesian}. In this study, we define input variables, including user, time, and location information, as $\{U, T, L\}$, represented by instances $\{u, t, l\}$. We denote intermediate variables $\{H, G\}$ as $\{h, g\}$. The output variable $Y$ represents the effect variable, which means predicted visiting locations. The relationship between these variables and their effects on outcomes $Y$ can be expressed by the following equation:
\begin{equation}\label{general output}
Y_{h, g, l} = Y(H=h, G=g, L=l),
\end{equation}
where the intermediate variables are $h = H_{u,t} = H(U=u, T=t)$ and $g = G_{h,l} = G(H=h, L=l)$. Variable $H$ is calculated with user and time information and is independent of $G$ and $L$ simultaneously; therefore, we denote it as $h$ to represent location-independent information in the following section. Then, if we want to know the causal effect of location information, we can calculate causal effect $E$ between different previous locations while predicting next locations:
\begin{equation}\label{probability gain}
E_{i, j} = Y_{h, g, l_{i}} - Y_{h, g, l_{j}}.
\end{equation}
\indent Next, we introduce \textbf{counterfactuals}, which address how the output variable $Y$ responds to values of input variables \cite{morgan2015counterfactuals}. This process involves $intervention$, which is executed using the $do(\cdot)$ operation. For instance, operation $do(G=g*)$ implies the creation of a counterfactual value $g*=G_{h*, l*}=G(H=h*, L=l*)$. Here, $g*$ is a hypothetical value that replaces the original value $g$ as it is impossible for us to get outcomes with unobserved input information. Under such an intervention, the output can be represented as follows:
\begin{equation}\label{invention}
Y_{h, g*, l} = Y(H=h, do(G=g*), L=l).
\end{equation}
\indent Fig. \ref{counterfactual} illustrates the mechanics of intervention in the counterfactual analysis. When a variable undergoes an intervention through the $do(\cdot)$ operation, it effectively severs all incoming links to the targeted node in the causal graph \cite{pearl2009causal}. However, this operation does not affect the other links in the graph. In the experimental setup, we use three strategies to construct a hypothetical value, denoted as $g*$, using manipulated inputs $h*$ and $l*$. Consequently, the output variable $Y$ is computed by considering three independent variables, namely $H$, the intervened $do(G=g*)$, and $L$.\\
\indent In general causal effects calculation, the focus is on understanding the total effects (TE), which pertain to how changes in input variables influence the output \cite{pearl2022direct}. The total effect is expressed using the following formula:
\begin{equation}
TE = Y_{h, g, l} - Y_{h*, g*, l*}.
\end{equation}
\indent Analysis of the causal graph depicted in Fig. \ref{proposed_causal} reveals that the total effect (TE) can be categorized into two distinct components: the natural direct effect (NDE) and the total indirect effect (TIE). In the proposed model, we identified two pathways each for NDE and TIE. The NDE, represented as $H \rightarrow Y$ and $L \rightarrow Y$, encapsulates the causal effect of input variables on the output when intermediate variables are subjected to intervention via the $do(\cdot)$ operation. The incoming links to the intermediate variable are severed during such an intervention. This is because the intervention effectively replaces the intermediate variable with a counterfactual value, nullifying the input variables' effect on it. Therefore, the intermediate variable no longer mediates the relationship between the input and the output. Consequently, the direct causal effect can be described as follows:
\begin{equation}
NDE = Y_{h, g*, l} - Y_{h*, g*, l*}.
\end{equation}
\indent Finally, we represent the total indirect effect using the following equation:
\begin{equation}\label{TIE}
TIE = TE - NDE =Y_{h, g, l} - Y_{h, g*, l}.
\end{equation}
\indent Total indirect effect (TIE) is crucial in predicting with nonanchor targeted travels, as depicted in Fig. \ref{nonanchor}. We treat them differently and compare them with anchor-targeted travels. The TIE is focused on the indirect effect through intermediate variables, removing the direct effect from inputs to outputs. We achieve this objective by introducing counterfactuals, which emphasize the role of intermediate variables in the proposed causal graph.
\subsection{Causality-Aware Predictor}\label{model}
In this section, we propose a general framework as Fig. \ref{framework} to implement causal inference into next location prediction with variables we define in section \ref{causality}. In order to provide a comprehensive explanation of our proposed method, we classify the currently dominant model structures into \textit{recurrent-based}, \textit{attention-based} and \textit{transformer-based}. Then, we implement counterfactual as an intervention to apply causal inference in next location predictions with these model structures.
\begin{figure}[htbp]
\begin{center}
\includegraphics[height=3in,scale=0.54]{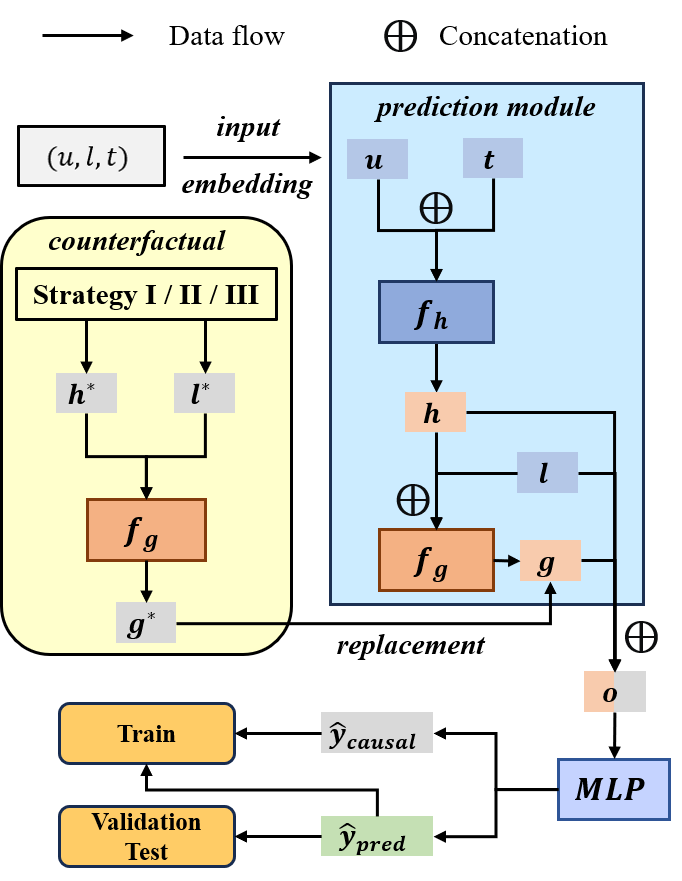}
\end{center}
\caption{Structure of causality-aware next location prediction framework. Counterfactuals will replace the original hidden state $\textbf{\textit{g}}$. Then, intervened prediction results with $\textbf{\textit{g*}}$ will be used for training by calculating $\hat{y}_{causal}$ but will not exist during validation and test processes.}
\label{framework}
\end{figure}\\
\textbf{Feature Embedding.} We first need to map a set of discrete inputs $\{u, l, t\}$ to a low-dimensional vector space to obtain their representations. In our task, we capture the continuous hidden states from the trajectories; therefore, we use embedding layers to encode user, location and time information, as introduced in Definition \ref{mobility data}. We denote these three embedded features as $\boldsymbol{u} = \boldsymbol{e}^{d}_{u}$, $\boldsymbol{l} = \boldsymbol{e}^{d}_{l}$, $\boldsymbol{t} = \boldsymbol{e}^{d}_{t}$, where $d$ is the dimension of the embedded features.\\
\textbf{Prediction Modules.} After embedding layers, prediction models need to process embedded input data with different structures. With three mentioned model structures to predict the next locations, we build two separated prediction modules $\textbf{\textit{f}}_{h}$ and $\textbf{\textit{f}}_{g}$. Fig. \ref{proposed_causal} and Fig. \ref{framework} display two hidden states with two separated modules. The first hidden state $\boldsymbol{h}$ is generated from user and time information, and the second hidden state $\boldsymbol{g}$ combines the first hidden state and location information. As we mainly focus on location information, we assign three types of prediction structures to $\textbf{\textit{f}}_{g}$ and for $\textbf{\textit{f}}_{h}$ we will fuse user and time information with simple operations like a multi-layer perceptron (MLP) or concatenation, although for implementation there are some modifications according to baseline. The details are as follows:
\begin{equation}
\boldsymbol{h}_{\tau} = \textbf{\textit{f}}_{h} (\boldsymbol{u}_{1:\tau} \oplus \boldsymbol{t}_{1:\tau}),
\end{equation}
\begin{equation}
\boldsymbol{g}_{\tau} = \textbf{\textit{f}}_{g} (\boldsymbol{h}_{1:\tau} \oplus \boldsymbol{l}_{1:\tau}),
\end{equation}
where $\tau$ is the length of the input trajectory, and $\oplus$ is the concatenation operation. Finally, we treat the last outputs $\boldsymbol{h}_{\tau}$ and $\boldsymbol{g}_{\tau}$ as intermediate products, as displayed in Fig. \ref{proposed_causal}, and use them to predict the next location.\\
\textbf{MLP Output Layer.} As displayed in Fig. \ref{framework}, we concatenate $\boldsymbol{h}_{\tau}$, $\boldsymbol{g}_{\tau}$ and $\boldsymbol{l}_{\tau}$ and send the fused features to a MLP layer to calculate the prediction results as follows:
\begin{equation}
\hat{y}_{pred} =  \boldsymbol{f}_{mlp}(\boldsymbol{h}_{\tau} \oplus \boldsymbol{g}_{\tau} \oplus \boldsymbol{l}_{\tau}).
\end{equation}
Generally, an MLP layer consists of several linear layers and activation functions. We use two linear layers and a $tanh$ activation function between them to construct the MLP function $\boldsymbol{f}_{mlp}$. Finally, we use $softmax$ function to calculate the probability of visiting each location, represented as $\hat{y}_{pred}$, and notably $\hat{y}_{pred}$ is output we mentioned in Equation \ref{general output}.\\
\textbf{Counterfactual Intervention.} This study aim to improve the prediction performance for nonanchor-targeted travels $\mathcal{T}^{2}$ confounded by direct effects after stratification. This intervention requires us to create counterfactual values $\boldsymbol{h}^{*}$ and $\boldsymbol{l}^{*}$ to calculate variable $\boldsymbol{g}^{*}$ to get $\hat{y}_{causal}$ as displayed in Fig. \ref{framework} and it represents the specific output in equation \ref{invention}. Simultaneously, a reasonable counterfactual assumption provides a correct perception of the effect of intermediate variables during the prediction process \cite{wang2023causal}. Therefore, we propose three strategies to generate counterfactual values:\\
\textbf{Strategy \uppercase\expandafter{\romannumeral1}}: In this strategy, we generate a random value with a uniform distribution: 
\begin{equation}
\boldsymbol{e}^{*} \sim U[0, 1).
\end{equation}
\textbf{Strategy \uppercase\expandafter{\romannumeral2}}: This strategy removes all information from original values with $\boldsymbol{\textbf{0}}$ and can be regarded as a special condition of strategy \uppercase\expandafter{\romannumeral1}.\\
\textbf{Strategy \uppercase\expandafter{\romannumeral3}}: This strategy is inspired by \cite{Chen_2021_ICCV}. The counterfactual asks a what if problem for an observation value, so we want to know, for an observation $e_{i}$, what if we replace it with other observations? Therefore, the counterfactual of the $i$th observation is the collection of all other observations generated using the following:
\begin{equation}
\boldsymbol{e}^{*}_{i}  = mean(\sum_{j \neq i}{\boldsymbol{e}_{j}}).
\end{equation}
\indent Finally, as we focus on location information throughout the prediction process, we apply the three strategies for location embedding $\boldsymbol{l}^{*}$. For $\boldsymbol{h}^{*}$, we applied strategies \uppercase\expandafter{\romannumeral1}. As we said, the output with counterfactual $\hat{y}_{causal}$ refers to $Y_{h, g*, l}$, which is calculated by only direct effects. Finally, $\hat{y}_{pred} - \hat{y}_{causal}$, referring to the TIE in equation \ref{TIE}, is the predicted result with causal inference for nonanchor-targeted travels in $\mathcal{T}^{2}$ \cite{wang2023causal}.
\subsection{Multi-Task Learning Target}
This study consider trajectories in $\mathcal{T}^{1}$ as anchor-targeted travels and $\mathcal{T}^{2}$ as nonanchor-targeted. Therefore, we proposed a prediction framework to predict the next location. By contrast, for nonanchor-targeted travels in $\mathcal{T}^{2}$, we apply causal inference with counterfactual as mentioned in Section \ref{causality} and we use the same model parameters for anchor-targeted travels. We combine these training tasks and obtain the following loss function $\mathcal{L}$:
\begin{equation}
\mathcal{L}= \sum_{i \in \mathcal{T}^{1}}\mathcal{L}_{ce}(\hat{y}_{pred}, y_{i}) + \sum_{i \in \mathcal{T}^{2}}\mathcal{L}_{ce}(\hat{y}_{pred} - \hat{y}_{causal}, y_{i}),
\end{equation}
where $y_{i}$ is the ground truth label and $L_{ce}$ is the cross-validation loss function.
\section{Experiment}
In this section, we will present the experiment's details and the associated analysis.

\begin{table*}[h]
\centering
\caption{Causality-aware next location prediction with state-of-the-arts models}
\renewcommand{\arraystretch}{1.1}
\begin{tabular}{l|ccc|ccc|ccc} 
\toprule
\multicolumn{1}{c|}{} & \multicolumn{3}{c|}{Foursquare\_TKY} & \multicolumn{3}{c|}{Foursquare\_NYK} & \multicolumn{3}{c}{Blogwatcher} \\ 
\hline
\multicolumn{1}{c|}{Locations} & \multicolumn{3}{c|}{\begin{tabular}[c]{@{}c@{}}61858\\\end{tabular}} & \multicolumn{3}{c|}{\begin{tabular}[c]{@{}c@{}}38333\\\end{tabular}} & \multicolumn{3}{c}{\begin{tabular}[c]{@{}c@{}}159513 \\\end{tabular}} \\
\multicolumn{1}{c|}{Users} & \multicolumn{3}{c|}{\begin{tabular}[c]{@{}c@{}}2293\\\end{tabular}} & \multicolumn{3}{c|}{\begin{tabular}[c]{@{}c@{}}1083\\\end{tabular}} & \multicolumn{3}{c}{\begin{tabular}[c]{@{}c@{}}4991\\\end{tabular}} \\ 
\multicolumn{1}{c|}{Records} & \multicolumn{3}{c|}{\begin{tabular}[c]{@{}c@{}}537703\\\end{tabular}} & \multicolumn{3}{c|}{\begin{tabular}[c]{@{}c@{}}227428 \\\end{tabular}} & \multicolumn{3}{c}{\begin{tabular}[c]{@{}c@{}}2859572\\\end{tabular}} \\ 
\hline
\multicolumn{1}{c|}{Metrics} & Recall@5 & MRR@5 & NDCG@5 & Recall@5 & MRR@5 & NDCG@5 & Recall@5 & MRR@5 & NDCG@5 \\ 
\hline
GRU                                             & 39.76 & 26.83 & 30.05 & 35.00 & 22.70 & 25.76 & 68.57 & 53.96 & 57.65 \\
+Strategy \uppercase\expandafter{\romannumeral1}& \textbf{44.42} & \textbf{30.88} & \textbf{34.26} & 41.33 & 28.01 & 31.34 & 71.05 & 56.29 & 60.02 \\
+Strategy \uppercase\expandafter{\romannumeral2}& 43.91 & 30.59 & 33.92 & \textbf{41.68} & \textbf{28.34} & \textbf{31.68} & \textbf{71.19} & \textbf{56.39} & \textbf{60.13} \\
+Strategy \uppercase\expandafter{\romannumeral3}& 44.19 & 30.76 & 34.11 & 41.27 & 27.92 & 31.26 & 71.08 & 56.33 & 60.06 \\
\hline
DeepMove                                        & 41.36 & 28.71 & 31.86 & 39.27 & 26.66 & 29.80 & 66.23 & 52.80 & 56.19 \\
+Strategy \uppercase\expandafter{\romannumeral1}& \textbf{44.66} & \textbf{31.26} & \textbf{34.60} & 44.28 & 30.54 & 33.96 & \textbf{67.72} & 53.98 & \textbf{57.46} \\
+Strategy \uppercase\expandafter{\romannumeral2}& 43.98 & 30.77 & 34.06 & 43.25 & 29.93 & 33.25 & 67.60 & 53.97 & 57.42 \\
+Strategy \uppercase\expandafter{\romannumeral3}& 44.28 & 31.04 & 34.34 & \textbf{44.41} & \textbf{30.69} & \textbf{34.11} & \textbf{67.62} & \textbf{53.99} & 57.44 \\
\hline
Flashback                                       & 36.82 & 25.11 & 28.03 & 34.37 & 22.30 & 25.30 & 70.01 & 54.69 & 58.55 \\
+Strategy \uppercase\expandafter{\romannumeral1}& \textbf{43.18} & \textbf{29.97} & \textbf{33.27} & \textbf{42.08} & \textbf{28.32} & \textbf{31.76} & 72.89 & \textbf{57.66} & 61.51 \\
+Strategy \uppercase\expandafter{\romannumeral2}& 43.11 & 29.93 & 33.22 & 41.39 & 27.94 & 31.30 & 72.92 & \textbf{57.66} & \textbf{61.52} \\
+Strategy \uppercase\expandafter{\romannumeral3}& 43.15 & \textbf{29.97} & 33.26 & 41.72 & 27.99 & 31.43 & \textbf{72.94} & 57.54 & 61.44 \\
\hline
HSTLSTM                                         & 39.08 & \textbf{27.07} & 30.06 & 37.38 & 25.08 & 28.15 & \textbf{80.24} & \textbf{61.23} & \textbf{65.99} \\
+Strategy \uppercase\expandafter{\romannumeral1}& 39.78 & 26.99 & 30.17 & 37.91 & 25.53 & 28.61 & 71.89 & 54.99 & 59.23 \\
+Strategy \uppercase\expandafter{\romannumeral2}& \textbf{39.86} & \textbf{27.07} & \textbf{30.25} & \textbf{38.12} & \textbf{25.67} & \textbf{28.77} & 78.03 & 59.49 & 64.14 \\
+Strategy \uppercase\expandafter{\romannumeral3}& 39.80 & 27.05 & 30.22 & 37.70 & 25.23 & 28.34 & 78.05 & 59.65 & 64.27 \\
\hline
GETNext                                         & 55.42 & 34.20 & 39.48 & 53.53 & 33.80 & 38.69 & 80.05 & 59.76 & 64.85  \\
+Strategy \uppercase\expandafter{\romannumeral1}& 55.81 & 35.14 & 40.30 & 55.08 & 34.53 & 39.63 & 80.05 & 59.79 & 64.86 \\
+Strategy \uppercase\expandafter{\romannumeral2}& 55.76 & 34.88 & 40.09 & \textbf{55.49} & \textbf{35.19} & \textbf{40.23} & 79.99 & 59.82 & 64.87 \\
+Strategy \uppercase\expandafter{\romannumeral3}& \textbf{55.95} & \textbf{35.26} & \textbf{40.42} & 54.55 & 33.90 & 39.03 & \textbf{80.16} & \textbf{59.88} & \textbf{64.96} \\
\bottomrule
\end{tabular}
\label{results}
\end{table*}
\subsection{Datasets and Preprocessing}
This study use real-life human mobility datasets, notably Foursquare and Blogwatcher, to test the effectiveness of the proposed causality-aware prediction framework. We provide an in-depth description of these datasets.:\\
\indent  \textbf{Foursquare check-in data:} This dataset comprises user check-in records from Tokyo and New York, collected using the Foursquare API \cite{yang2014modeling}. Each record includes a user ID, timestamp, GPS location, and POI ID. For preprocessing, we arrange the check-in records for each user chronologically. Trajectories are segmented using a 72-hour temporal interval, and we filter trajectories with fewer than five records and users with fewer than five trajectories. \\
\indent  \textbf{Blogwatcher Data:} Provided by Blogwatcher Inc, this private dataset contains GPS records from users in Japan, collected during October and November 2022. For our experiment, we focus on records from Japan's Kanto region (including the Tokyo, Chiba, Kanagawa, and Saitama Prefectures). The GPS data are aggregated into Japan's 125-meter grid indexes as location IDs. We filter out user IDs with more than 500 records after stay-point detection to align Foursquare check-in datasets \cite{pappalardo2019scikit}. We preprocess the filtered dataset with settings similar to those used in Foursquare; however, for the time window to cut trajectories, we chose 2 hours as the stay point detection with GPS data generated much more locations than active check-in behaviors did.
\subsection{Experimental Settings}
The dataset is divided into training, validation, and test sets with a ratio of 0.7/0.1/0.2. The best hyperparameters are determined based on the validation set using k-fold cross-validation. Each test utilizes a uniform model structure, maintaining the same number of parameters. In conventional models, two travel patterns are merged together for training, whereas for causality-aware prediction model, nonanchor targeted travels are implemented with counterfactuals as causal inference for training as we illustrated in Fig. \ref{framework}. Embedding dimensions are set to 128 for the user, location, and time information and 256 for the final hidden state if the original paper do not provide detailed information. The experiments are conducted on a dl-box GPU server with four NVIDIA RTX A6000 graphic cards, using Python 3.9, PyTorch 1.9, and Cuda Toolkit 11.7.
\subsection{Performance Analysis}
To compare performance of our proposed framework, we select several state-of-the-art baseline methods as follows:
\begin{itemize}[topsep=0pt]
    \item \textbf{GRU} \cite{chung2014empirical}: a basic recurrent-based module, widely used in the temporal sequential data prediction task.
    \item \textbf{DeepMove} \cite{feng2018deepmove}: a state-of-the-art model with attention-based recurrent neural network that extracts periodical patterns from historical trajectories.
    \item \textbf{Flashback} \cite{yang2020location}: a state-of-the-art model with a context-aware hidden state weighting mechanism generated from spatial and temporal information.
    \item \textbf{HSTLSTM} \cite{kong2018hst}: a state-of-the-art model that combines the LSTM module with the spatiotemporal factor calculated with physical distances.
    \item \textbf{GETNext} \cite{yang2022getnext}: a transformer-based state-of-the-art model with graph-based representation learning module and semantic information of locations.
\end{itemize}
\indent As mentioned in Section \ref{formulation}, we choose $Recall@k$, $MRR@k$, and $NDCG@k$ as metrics to compare the performance of the different models, where $K$ is the top prediction output of $K$ during the test process.
\indent Table \ref{results} presents the performance of the baseline models using various counterfactual strategies for next location prediction. The results reveal that the proposed module considerably improves the performance of the baseline models under most conditions, highlighting the efficiency of causality analysis in stratified human mobility data. We evaluated prediction performance using the Foursquare and Blogwatcher datasets, which represent human mobility in active check-in behaviors and passive stay-point detection, respectively. The results listed in Table \ref{results}, indicate improvements in prediction accuracy as $Recall@5$: $11.70\%$, $7.98\%$, $17.27\%$, $1.99\%$ and $0.96\%$ in the Foursquare Tokyo dataset with four baseline models; $19.09\%$, $13.09\%$, $20.63\%$, $1.98\%$ and $3.66\%$ in the Foursquare New York dataset; finally, in the Blogwatcher data, the proposed model improved the baseline model performance by $3.79\%$, $0.74\%$, $4.19\%$, $-2.73\%$ and $0.14\%$. Our proposed model performs better in most conditions, providing a novel perspective on causality analysis for next location prediction. 
\indent For counterfactual strategies, we create a hypothetical value to replace the real value. Strategies \uppercase\expandafter{\romannumeral1} and \uppercase\expandafter{\romannumeral2} generate superior prediction performance, whereas Strategy \uppercase\expandafter{\romannumeral3} is only effective under certain conditions. This analysis, coupled with the minus operation ($-$) used to calculate the TIE in deep learning, encourages the model to output a higher possibility of correct answers and minimizes the randomness of the predictions \cite{niu2021counterfactual,wang2023causal}.
\subsection{Hyperparameter Analysis}\label{hyper}
\begin{figure}[ht]
\begin{center}
\includegraphics[height=1.4in,scale=0.39]{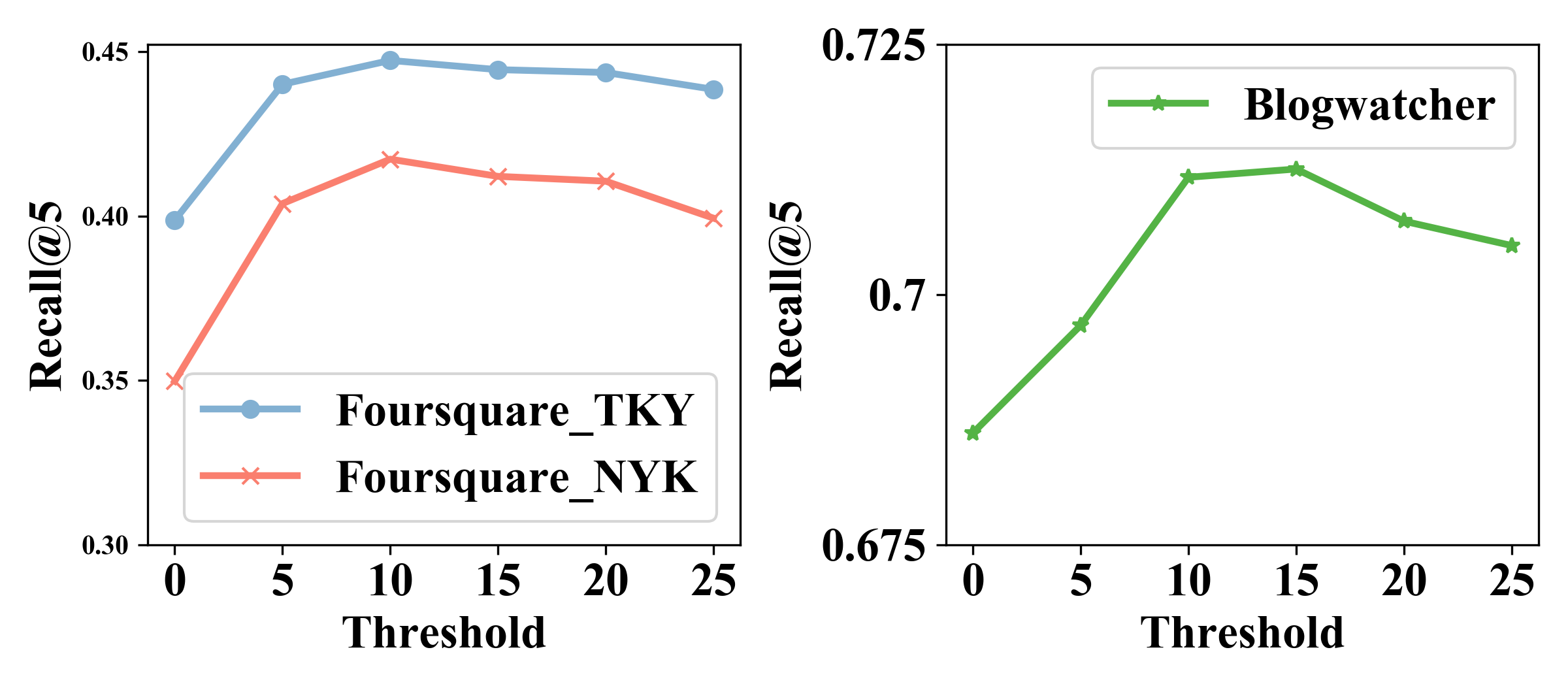}
\end{center}
\caption{Hyperparameter analysis of the anchor threshold.}
\label{threshold}
\end{figure}
\indent Consistent with Section \ref{related} and Definition \ref{anchor locations} and \ref{stratification} in Section \ref{formulation}, we stratify human mobility data based on the predefined frequency threshold by individual visits to each destination to obtain anchor- and nonanchor-targeted travels. As depicted in Fig. \ref{threshold}, we tested various threshold candidates in $\{0, 5, 10, 15, 20, 25\}$ to evaluate the effectiveness of our causality-aware next location prediction model. For simplicity in preprocessing, if the visit frequency of an individual to a particular location exceeds a fixed threshold, all travels to that location are designated as anchor targeted. The experiments using the GRU in the baseline models are evaluated using the Foursquare and Blogwatcher datasets. The results indicate that a threshold of $10$ for Foursquare and $15$ for Blogwatcher can optimally stratify human mobility data into two groups: anchor- and nonanchor-targeted groups. When we do not implement causality analysis on all travels, that is, the threshold was set as $0$, the model achieves a performance similar to a conventional model, regardless of whether the causal inference module is activated. However, setting a higher threshold results in a performance decline because of fewer destinations classified as anchor locations, which increases the difficulty of the model to learn regular travel patterns from daily periodic behaviors.

\subsection{Ablation Study}
\begin{table}[h]
\centering
\renewcommand{\arraystretch}{1.1}
\caption{Ablation study based on proposed causal graph}
\begin{tabular}{l|ccc} 
\toprule
\multicolumn{1}{c|}{} & \multicolumn{3}{c}{Foursquare\_TKY}\\ 
\hline
\multicolumn{1}{c|}{Metrics} & Recall@5 & MRR@5 & NDCG@5 \\ 
\hline
ours (best)                                                          &\textbf{44.42} & \textbf{30.88} & \textbf{34.26}\\
w/o \textit{link 1 ($H \rightarrow Y$)}                              & 43.90 & 30.44 & 33.80\\
w/o \textit{link 2 ($L \rightarrow Y$)}                              & 41.44 & 28.46 & 31.70\\
w/o \textit{link 1 \& link 2}                    & 41.10 & 28.37 & 31.55\\
\bottomrule
\end{tabular}
\label{ablation}
\end{table}
To validate the effectiveness of the proposed causal graph illustrated in Fig. \ref{proposed_causal}, we perform an ablation study by cutting links in the graph. In this section, we evaluate the necessity of novel proposed links in Fig. \ref{proposed_causal} and remove the direct causal effects of $H \rightarrow Y$, represented as $link \ 1$ and $L \rightarrow Y$ represented as $link \ 2$, in the GRU baseline model with Foursquare Tokyo as the test dataset. Table \ref{ablation} presents the ablation study results, and which reveal that the two edges of direct causal effects are indispensable for improving prediction accuracy. By removing two links, the performance drop by $7.47\%$, but it is superior to that of the conventional models because we implement causal inference in our test. Moreover, by removing $link \ 1$ and $link \ 2$, model performance decline by $1.17\%$ and $6.71\%$, respectively. Throughout the ablation study, $link \ 2$ contribute to the prediction performance than $link \ 1$ does; therefore, $link \ 2$, which connects the previous location with the next location, represents spatially continuous dependencies.

\subsection{Qualitative Study}
\begin{figure}[ht]
\begin{center}
\includegraphics[height=2.4in,scale=0.42]{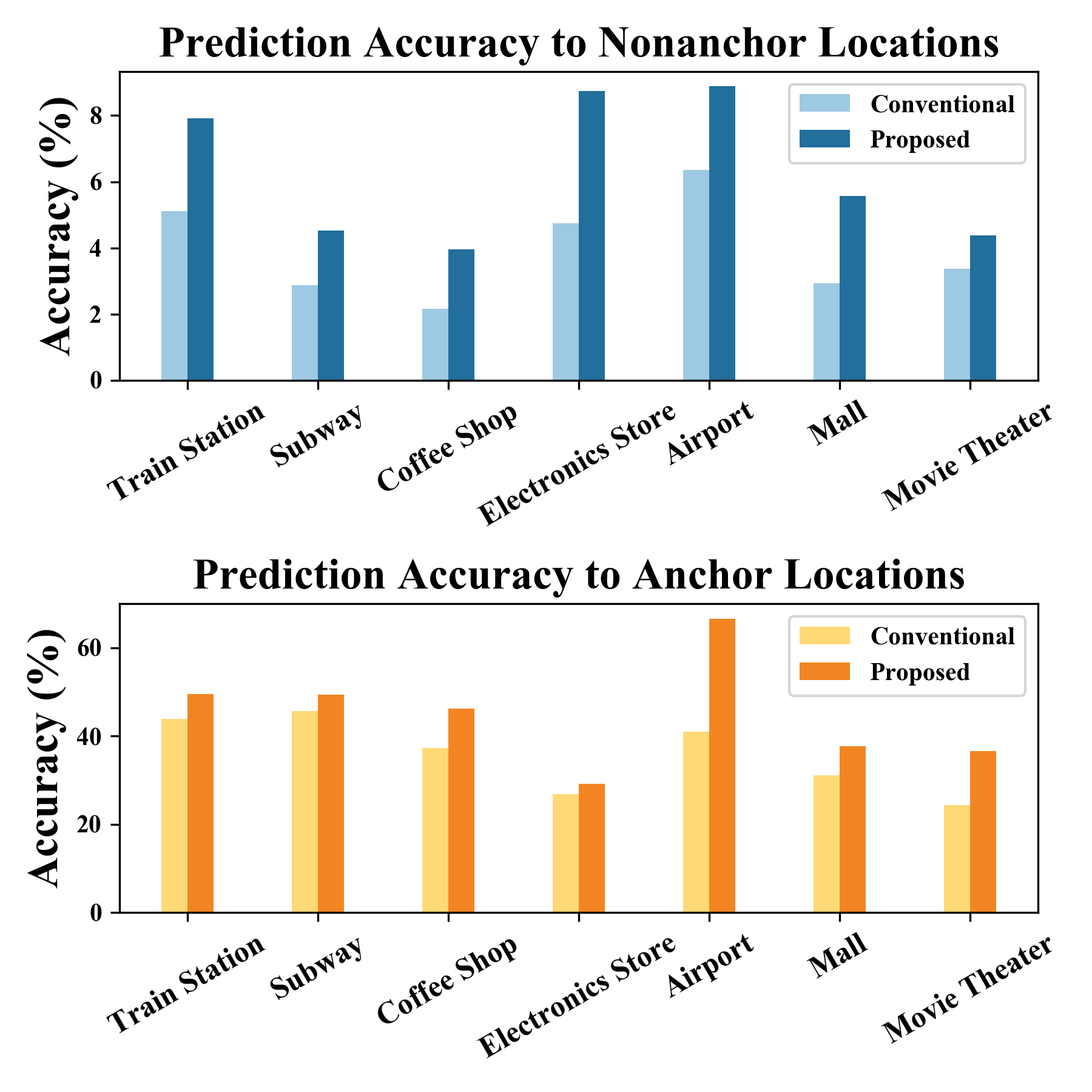}
\end{center}
\caption{Comparison of conventional and proposed framework for multiple categories in Foursquare Tokyo.}
\label{comparison}
\end{figure}
We analyze prediction results in various categories in the Foursquare Tokyo test dataset to assess the effectiveness of the proposed model. This evaluation is conducted using both the conventional GRU model and the proposed causality-aware GRU model with strategy \uppercase\expandafter{\romannumeral1}. We focus on several prevalent POI categories in the dataset. In particular, these POIs can be considered anchor and nonanchor locations because we categorized anchor locations at the individual level. For example, although $Train \ Station$ can be an anchor location for some users who frequently check-in at stations, it may not be as significant for others. Fig. \ref{comparison} shows the prediction performance for each POI type within the dataset, highlighting the variance in magnitude. Fig. \ref{comparison} shows that the proposed model improved the prediction accuracy for anchor and nonanchor locations. This improvement is notable for nonanchor locations compared to conventional prediction methods. Consequently, the model not only addresses the problem of previous location information interfering with predictions in conventional models for anchor locations, as we discussed before, it considerably boosts the predictive accuracy for nonanchor locations within a unified model. This result reveals that causality analyses are necessary as different travel patterns are fused into human mobility data.
\section{Conclusion}
In this study, we develop a causality-aware next location prediction framework with human mobility stratification. Unlike conventional methods, we create a novel causal graph to indicate causal effects among variables during prediction, such as \textit{user}, \textit{location}, \textit{time} and hidden states generated by the models. The proposed framework is evaluated using several state-of-the-art models accompanied by three counterfactual strategies to remove direct effects during prediction. The integration of causal inference into our model results in a notable performance improvement across two human mobility datasets, Foursquare and Blogwatcher. This improvement remains robust in most conditions, highlighting the effectiveness of the proposed method in improving the accuracy and reliability of the next location prediction.

\bibliographystyle{IEEEtran}
\bibliography{IEEEexample}

\end{document}